\documentclass[12pt]{article}

\usepackage{framed}
\usepackage{newunicodechar}
\usepackage[margin=1in]{geometry}
\usepackage{setspace}
\usepackage{xcolor}
\usepackage[normalem]{ulem} 
\usepackage{graphicx}
\usepackage[section]{placeins}
\usepackage{booktabs}
\usepackage{tabularx}
\usepackage{enumitem}
\usepackage[hidelinks]{hyperref}
\usepackage[utf8]{inputenc}

\usepackage{biblatex}
\newcommand{\subheading}[1]{\vspace{0.75em}\noindent\textbf{#1}\par}

\newenvironment{ticsbox}[1]
{\par\medskip\begin{framed}\noindent\textbf{#1}\medskip\par\noindent}
{\end{framed}\par\medskip}

\title{The Cartesian Cut in Agentic AI}
\author{
\textbf{Tim Sainburg}\\
Harvard University
\and
\textbf{Caleb Weinreb}\\
Harvard Medical School
}\date{}

\begin{document}

\maketitle
\thispagestyle{plain}


\section*{Highlights}
\begin{itemize}[leftmargin=1.2em]
\item Brains are fundamentally optimized for layered feedback control of embodied behavior. Prediction in brains is a byproduct of control.
\item Large language model (LLM) based agents are first optimized for text prediction and then retrofitted for control through tools and orchestration.
\item The grafting of a predictive core to an engineered runtime is referred to as "Cartesian agency."
\item Cartesian agency enables modular tooling and governance, while introducing a control-integration bottleneck across the model/runtime boundary.
\item We use the concept of Cartesian agency to distinguish a spectrum of agent design pathways that make different trade-offs between autonomy, robustness, and oversight.
\end{itemize}

\section*{Abstract} 
LLMs gain competence by predicting words in human text, which often reflects how people perform tasks. Consequently, coupling an LLM to an engineered runtime turns prediction into control: outputs trigger interventions that enact goal-oriented behavior. We argue that a central design lever is where control resides in these systems. Brains embed prediction within layered feedback controllers calibrated by the consequences of action. By contrast, LLM agents implement Cartesian agency: a learned core coupled to an engineered runtime via a symbolic interface that externalizes control state and policies. The split enables bootstrapping, modularity, and governance, but can induce sensitivity and bottlenecks. We outline bounded services, Cartesian agents, and integrated agents as contrasting approaches to control that trade off autonomy, robustness, and oversight.


\section*{A Cartesian split in agentic AI}
LLMs have shown that broad competence can emerge from predictive training objectives when scaled in data and compute \cite{brown2020language, kaplan2020scaling, radford2019language, bommasani2021foundation}. When these models are embedded in agents that convert text into tool calls, environment queries, or code execution, their ability to mimic human-like reasoning enables the composite systems to perform human-like tasks. It is possible that this recipe for artificial agency (trace-first predictive pretraining followed by post-training, tool use, and orchestration) will scale into robust, long-horizon general intelligence \cite{kaplan2020scaling, wei2022emergent, bubeck2023sparks, chen2026_doesAI}. An opposing view emphasizes the contrast between such agents and biological brains, in which cognition is constitutively coupled to action, exploration, and the consequences of intervention \cite{pezzulo2024passiveai, bender-koller-2020-climbing}. Although brains do predict, that is not what they evolved for; prediction is merely one faculty among many that enables the regulation of behavior under feedback \cite{cisek2007cortical, cisek2022evolution, todorov2002optimal}. More importantly, the parts of the brain that do the predicting are not functionally separated from the parts that control muscles, store memories and process sensory feedback.

It remains unclear which aspects of brain architecture are necessary for the human-like abilities that artificial agents still lack, since much of brain organization reflects contingent evolutionary contraints such as metabolic resources, developmental history, and homeostatic regulation, rather than minimal requirements for general intelligence per se \cite{cisek2022evolution, bostrom2014superintelligence, legg2007universal}. Thus, the purpose of this manuscript is not to argue that artificial agents must become more brain-like, nor to propose a single optimal architecture \cite{tomasello2025howto, pezzulo2024passiveai}. Rather, the contrast between biological and artificial agents highlights a particular design choice in how artificial agents are built. The same LLM can yield reliable behavior or be surprisingly unstable depending on wrapper choices, including prompt and schema conventions, memory serialization, tool routing, and stopping/retry logic. Architecturally, this reflects the boundary between a predictive core that emits symbolic traces (plans, rationales, tool calls) and an external runtime that turns those traces into plans, memories, and interventions by executing tools and enforcing permissions, recovery, and fallback policies \cite{yao2022react, schick2023toolformer, chase2022langchain}. 
We call this decomposition ``Cartesian agency,'' by analogy to Descartes' mind--body split: not as a metaphysical claim, but as a label for a separation within the control stack. The boundary is the ``Cartesian cut'' that induces a ``Cartesian split'' between a learned model (formed primarily from predictive training objectives) and engineered control (encoded in wrappers, tools, and policies). {This dissociation also gives the debate over the ``extended mind'' an unusual twist: in Cartesian agents, externalization is not confined to peripheral aids such as notebooks or calculators, but can encompass functions often treated as core to cognition itself, including memory, action selection, and the interface to sensing and actuation.} Although clearly distinct from human agency \cite{pezzulo2024passiveai}, the Cartiesian split is not intrinsically misguided.
The cut is a genuine source of leverage: it bootstraps competence from traces of human behavior (encoded in text corpora) and yields modular, instrumentable interfaces. On the other hand, it forces control-relevant state through a narrow symbolic protocol, limiting the system's cognitive capacity and making it brittle in the face of small changes to the control interface \cite{lightman2023let}.

From this perspective, current efforts to engineer more capable agents can be organized into different pathways that strengthen or weaken the Cartesian cut. One pathway emphasizes bounded services (or boxed cognition): systems that provide planning, forecasting, verification, and synthesis while remaining embedded in human control loops (e.g., Comprehensive AI Services \cite{drexler2019cais}, Scientist AI \cite{bengio2025superintelligent}). An alternative pathway pursues integrated agents that internalize more of the control stack end-to-end, motivated by the idea that robust autonomy may require agents that tightly couple perception, action, memory and decision-making; learning directly from action and feedback, representing uncertainty about the consequences of intervention, and learning when to plan, act, or seek information over different timescales \cite{pezzulo2024passiveai, schrittwieser2020mastering, hafner2020mastering, reed2022generalist, zitkovich2023rt}. Each pathway carries characteristic advantages and liabilities, trading off autonomy and robustness on the one hand with transparency and steerability on the other. The remainder of the paper develops this framing by (i) grounding a biological baseline of integrated feedback control, (ii) analyzing Cartesian agency as an explicit design pattern that inverts biological control, and (iii) exploring the types of tasks for which Cartesian agents are well-suited versus those which may require greater integration.

\begin{figure}[t]
  \centering
  \includegraphics[width=\linewidth]{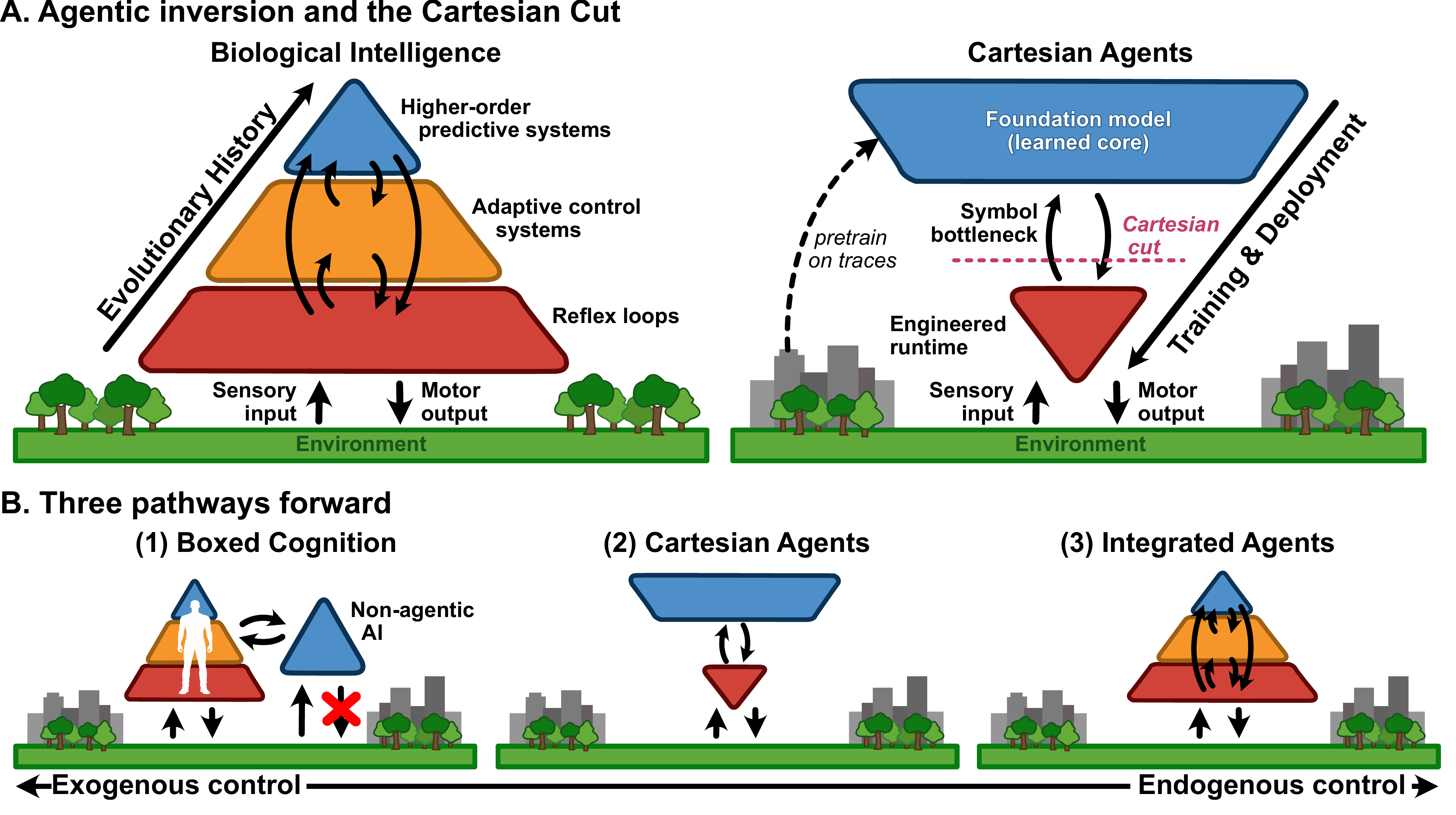}
  \caption{
  \protect Agentic inversion, the Cartesian cut, and pathways for relocating control.  \textbf{(A)} Biological intelligence (left) can be viewed as an evolutionary stack of nested feedback controllers: fast reflex loops support adaptive control systems, which in turn support higher-order predictive systems; sensory inputs and motor outputs close the loop with the environment. Thus, control is not top-down. Cartesian Agents (right) begin with predictive pre-training on passive traces (a learned core) and achieve control at deployment by coupling this core to an engineered runtime (executor/orchestration layer) that manages tools, policies, and execution. The interface between learned core and engineered runtime is the Cartesian cut, creating a low-bandwidth \emph{symbol bottleneck} through which control-relevant information must pass. The inverted geometry is schematic and intended to emphasize this order of construction (control-first in biology versus prediction-first in many engineered systems).
  \textbf{(B)} Three pathways forward organized by where control resides: boxed cognition keeps control predominantly exogenous (human/institutional loops; limited actuation), Cartesian agents implement mixed control via a learned core plus exogenous orchestration, and integrated agents shift more control endogenously by learning arbitration, memory, and adaptation within the agent. The horizontal axis schematizes a shift from exogenous to endogenous control.}
  \label{fig:cartesian_cut}
\end{figure}

\section*{The brain is a web of control}

Nervous systems evolved to regulate behavior under feedback \cite{Powers1973Behavior,cisek2007cortical}. The brain's layered architecture reflects this evolutionary history: newer circuits rarely replace older controllers outright; instead they modulate, bias, and coordinate them \cite{cisek2022evolution} (Fig. \ref{fig:cartesian_cut}A left). Indeed, this layering of feedback regulation predates nervous systems entirely; even single-celled organisms implement closed-loop policies that couple sensing to action. For example, \textit{E.~coli} uses chemotaxis to bias movement up chemical gradients, increasing exposure to favorable environments \cite{keegstra2022ecological}. As nervous systems evolved, layers of improved sensory representations, motor plans, state estimation, and prediction, allowed organisms to choose actions that are more robust to delay, perturbations, and partial observability. For example, fast spinal and brainstem loops stabilize posture, breathing, and locomotion; higher loops arbitrate among competing behaviors and shape learning through reinforcement and motivation; and still higher representations coordinate behavior across contexts and longer timescales (Box 1). This distributed architecture allows each neural system to focus on a particular level of behavioral organization. For example, neurons in the spinal cord can focus on muscles and proprioceptors without worrying about where the animal is going and why. As such, each system can maintain a distinct internal state, implement its own recurrent dynamics, and learn with domain-appropriate plasticity rules. This is quite unlike the architecture of Cartesian agents in which all internal states and thus recurrent dynamics must pass through a narrow symbolic bottleneck (i.e. become formatted as tokens and stored in context). 

LLMs and brains both make predictions. Yet the function and origin of these predictions are profoundly different.
In brains, the primary function of prediction is to improve control by anticipating the consequences of candidate actions \cite{murray2014hierarchy,mante2013context}. For example, predictions in neocortex are thought to shape behavior by providing richer beliefs about latent state, expected outcomes, and context which are signaled through recurrent cortical-subcortical loops. Competing theoretical frameworks, including predictive coding, active inference, optimal feedback control, and affordance competition, disagree on mechanisms and emphasis, but they agree that neural representations -- including predictions -- emerge and are maintained only insofar as they improve the regulation of action under feedback \cite{rao1999predictive,friston2005theory,bastos2012canonical,todorov2002optimal,cisek2007cortical}. This contrasts with LLMs, where prediction is initially treated as an end in itself (during autoregressive pretraining) and only later co-opted to generate action.

\begin{ticsbox}{Box 1. Layers of biological control}
\textbf{Fast sensorimotor regulation.} Spinal and brainstem circuits implement rapid error-correcting reflexes and central pattern generators that stabilize posture, breathing, and locomotion \cite{GrillnerWallen1985CPG, kandel}. These loops define the high-bandwidth interface through which slower systems influence the body.

\vspace{0.35em}
\textbf{Action selection and gating.} Cortico--basal ganglia--thalamic loops help arbitrate among competing actions, gate learning and working memory, and allocate processing under uncertainty \cite{alexander1986parallel, Redgrave1999BasalGanglia}. 

\vspace{0.35em}
\textbf{Calibration and predictive correction.} Cerebellar circuits support error-driven calibration and forward-model-like prediction that increase robustness despite delay and noise \cite{wolpert1998internal}. 

\vspace{0.35em}
\textbf{Global learning signals.} Neuromodulatory systems broadcast low-dimensional signals (e.g., reward prediction errors) that tune learning rates, motivation, exploration, and prioritization across many subsystems \cite{schultz1997neural}.

\vspace{0.35em}
\textbf{Cortex as flexible predictive bias.} Expanded cortex supports high-dimensional, context-sensitive predictive representations that bias downstream control, support counterfactual evaluation, and enable flexible planning, while remaining embedded in older loops rather than replacing them \cite{cisek2007cortical, rao1999predictive, bastos2012canonical}.
\end{ticsbox}

\section*{Cartesian agency as a design pattern}

The biological baseline emphasizes a functional constraint: prediction acts in the service of intervention. Many contemporary LLM agents invert this relationship in deployment. They begin with a predictor trained predominantly on passive traces (text, code, images, and other records of human activity), and then retrofit control by surrounding the predictor with an engineered action interface and an orchestration layer. In practice, the resulting system is not a single monolithic controller, but a composite architecture whose behavior depends on a boundary between a learned predictive core and an engineered control substrate \cite{yao2022react, schick2023toolformer, nakano2021webgpt} (Fig. \ref{fig:cartesian_cut}A, right).

We use Cartesian agency to name this recurring architectural pattern. The term is not a claim about minds or metaphysics. It is an architectural description of how control is partitioned in an agent stack: a learned core produces symbolic traces, while an engineered runtime (policies, tools, permissions, and actuation code) turns those traces into real-world effects. Thus, the separation pushes beyond the “mind”/“body” metaphor: it is a split within the control stack itself that, in humans, would separate subsystems that normally remain tightly integrated. The strength of the Cartesian cut varies across systems, but its defining feature is that control-relevant state (e.g., tool schemas, stopping criteria, retry policies, memory serialization, and guardrails) is specified and enforced outside the learned model, in the orchestration/runtime layer. When the model needs to condition on this state, it is supplied only through the interface protocol (prompt text, JSON, or function-call tokens) (Box 2).

\begin{ticsbox}{Box 2. Anatomy of a Cartesian agent}
\textbf{Predictive core (learned core)}. A foundation model, typically pretrained on passive traces and further tuned (e.g., with supervision or RL), learns broad regularities over language and other symbolic records. It emits symbolic traces at inference time: rationales, plans, tool selections, and structured arguments.

\vspace{0.35em}
\textbf{Orchestration layer (controller).} A runtime constructs prompts, maintains state (memory, scratchpads, retrieved documents), and implements control policies: termination, retries, tool allowlists, sandboxing, rate limits, and routing among tools or submodels.

\vspace{0.35em}
\textbf{Actuation via tools (execution layer).} Tools execute computations and interventions (e.g., code execution, search, database queries, robotics skills). Tool outputs return as observations appended to context, closing a Thought--Action--Observation loop.

\vspace{0.35em}
\textbf{The Cartesian cut.} The interface between learned core and runtime is a constrained symbolic protocol (text/JSON/function calls) that the runtime can parse. In a strong Cartesian design, key control variables (e.g., permissions, stopping/retry logic, and memory serialization) are implemented in the runtime and become available to the core only when explicitly serialized into this protocol. Changing this protocol (schemas, prompt formatting, memory representation) can materially change behavior because it changes how control is externalized and communicated \cite{sclar2023promptformat, liu2024lostinthemiddle}.

\vspace{0.35em}
\textbf{Training is orthogonal to the cut.} The Cartesian cut is an inference-time architectural boundary: it is present whenever tool policies, memory formats, retry/termination logic, and other control variables are implemented in an external runtime and made available to the model only via explicit serialization. A core trained primarily by next-token prediction, by reinforcement learning from interaction, or by any mixture can therefore instantiate a Cartesian agent if this boundary remains intact.

\end{ticsbox}

A useful archetype is the loop popularized by tool-using prompting schemes such as ReAct \cite{yao2022react}. A user goal arrives as text; the runtime composes a prompt that includes tool descriptions, state, and prior observations; the model produces interleaved reasoning and an action specification; the runtime parses that action, executes a tool (e.g., a search API or code execution), and returns the result as an observation; the cycle repeats until a stopping condition is met. 
Crucially, many control primitives that determine real-world behavior---what tools exist, how they are called, what constitutes success, when to halt, how to recover from errors, and what persistent memory is available---are implemented outside the learned core. The model exerts leverage through the interface language, but the orchestration layer determines how that language is converted into interventions.

Cartesian agency works because it exploits a specific asymmetry between trace learning and embodied discovery: human traces are already the products of control. Text, code, and other records concentrate solutions, conventions, and error-corrections that would be expensive for an agent to rediscover by exploration. A trace-trained model therefore starts with powerful priors over symbolic action spaces: how to decompose tasks, how to use institutional procedures, and how to express intermediate states in interpretable formats.

The cut also enables modular cognitive tooling. Tools offload computation, search, and verification to external systems with well-specified, often testable outputs. For example, when the model expresses solutions as executable code and a runtime carries out that code, problem decomposition is handled by the model while execution and correctness are delegated to a reliable external runtime \cite{gao2022pal, chen2022pot}. The same separation appears in browser-assisted question answering, where the model interacts with an external information environment and its claims can be checked against retrieved sources \cite{nakano2021webgpt}. 

The same symbolic interface that makes these interactions explicit and parseable also makes them amenable to measurement and control: developers can log trajectories, constrain tool access, sandbox execution, impose rate limits, and exert control through automated checks and runtime monitoring. This makes Cartesian systems attractive for deployment, because many control and safety properties can be adjusted in interfaces and symbol bottlenecks rather than by training.

However, these advantages come with a structural cost. The learned core can only influence the world through a discrete symbolic protocol that must remain interpretable to the static runtime. This symbol bottleneck restricts the bandwidth by which higher-level modeling can shape lower-level control. This contrasts with biological systems where prediction, action selection, and calibration signals are coupled through dense recurrent dynamics: in the brain, decision-making is densely coupled to internal state and calibration signals (e.g. interoceptive variables, neuromodulatory gain/arousal, and sensorimotor prediction errors) that are continuously available to the control system rather than being re-encoded into an interpretive protocol \cite{craig2002interoception, astonjones2005locus, yudayan2005uncertainty, wolpert1998internal}. Conversely, in Cartesian agents, many control-relevant degrees of freedom (e.g., how memory updates under uncertainty, how confidence is represented under attentional limits, how alternative actions are evaluated under competing costs) are expressed only when they are translated into language or tool arguments. These variables can be approximated by engineering additional tools and state representations \cite{kadavath2022mostly,nakano2021webgpt,lightman2023let}, but this requires that human designers understand which variables matter for stable feedback regulation and how to expose them, a challenge that remains only partially understood in the science of neural and cognitive control.

Several familiar liabilities follow from this displacement. First, wrapper sensitivity arises because runtime control state is communicated through prompt templates, schemas, parsing conventions, and serialized memory. Seemingly superficial interface changes can materially change behavior even when task semantics are preserved \cite{sclar2023promptformat, liu2024lostinthemiddle}. This sensitivity can result in stifled capactities, but more insidiously it can hide a capability overhang \cite{ngo2024agencyoverhang}: a model may appear less capable than it is because of poorly integrated tooling, scaffolding, or interface design. Seemingly insignificant fixes to runtime then unlock large capability gains, making the effects of deployment modifications harder to anticipate and evaluate, introducing a safety risk. Second, the bottleneck can look more legible than it is. The same channel that carries actions and tool calls also carries natural-language rationales, but chain-of-thought traces are not guaranteed to be faithful explanations of the computations that drive outputs. They can be plausible post hoc rationalizations, and their content can be systematically manipulated without corresponding changes in the underlying decision basis \cite{turpin2023unfaithful, barez2025cotnotexplainability, greenblatt2024alignment}. Another consequence is limited calibration under intervention. Training on passive traces can produce agents that speak fluently about policies but have weakly grounded estimates of feasibility, uncertainty, and recovery when they act through a specific actuator in a specific environment. Without feedback from real consequences, their behavior may not update appropriately under intervention. For example, in sequential decision making, behavioral cloning from static logs is vulnerable to compounding error and policy-induced distribution shift \cite{ross2011dagger, levine2020offline}. Post-training and on-policy fine-tuning can mitigate these effects \cite{christiano2017preferences, ouyang2022instructgpt}, but they do not remove the structural fact that many control variables remain exogenous, partially observable, and must be communicated through an explicit interface rather than arising from tightly coupled internal dynamics, posing limits to the information the model can attain from exploration. 

Taken together, Cartesian agency is neither a mistake nor a guarantee of robustness; it is a design decision. Strengthening the cut by adding more tools, more verification, and tighter runtime policies can improve reliability and oversight in tool-mediated domains, but it can also increase wrapper dependence and amplify unstable interface sensitivities. Dissolving the cut by internalizing more arbitration, memory, and adaptation into learned control may improve robustness in feedback-rich settings, but it also increases autonomy and shifts oversight burdens inward. This motivates treating the locus of control as an explicit variable in how we reason about capability and safety. In the next section, we use it to organize three pathways in agent design (bounded services, Cartesian agents, and integrated agents) and articulate predictions for where each pathway should excel as environments demand tighter intervention calibration, longer horizons, and greater resilience under perturbation.

\section*{Three pathways: relocating control across the Cartesian cut}

The preceding section treated Cartesian agency as a recurrent design pattern: a learned predictive core coupled to an engineered control substrate that manages tool access, memory formats, retries, and termination. In current work on agentic AI, the most consequential architectural moves can be understood as two flanking responses to the Cartesian cut: either (i) keep the cut strong and move control outward into human institutions and explicit governance (bounded services / boxed cognition), or (ii) dissolve the cut and move control inward by learning more of the control stack end-to-end (integrated agents) (Table 1).

We can describe these alternatives in terms of exogenous versus endogenous control, where control denotes the mechanisms that couple prediction to action through feedback, arbitration, and correction over time. Exogenous control refers to control-relevant functions implemented outside the learned model: goal setting, permissions, memory serialization, stopping criteria, recovery policies, and verification. Endogenous control refers to those same functions being implemented inside the learned system through internal state, learned arbitration across timescales, and ongoing adaptation from the consequences of action. The three pathways below differ mainly in where they place these functions, and therefore in what kinds of robustness and oversight they can plausibly deliver.

\begin{table}[t]
\centering
\footnotesize
\caption{Three pathways organized around the Cartesian baseline (Pathway 2), distinguished by where control resides relative to the Cartesian cut.}
\label{tab:pathways}
\begin{tabularx}{\linewidth}{
  @{}
  >{\raggedright\arraybackslash}p{3.1cm}
  >{\raggedright\arraybackslash}p{3.0cm}
  >{\raggedright\arraybackslash}X
  >{\raggedright\arraybackslash}X
  @{}
}
\toprule
\textbf{Pathway} & \textbf{Control locus} & \textbf{Capability upside} & \textbf{Primary liabilities} \\
\midrule
(1) Bounded services / boxed cognition &
Predominantly exogenous (humans and institutions; tight runtime constraints) &
Useful for planning, verification, synthesis, and monitoring without persistent autonomous actuation &
Risk of control leakage through persuasion, miscalibrated advice, or human overreliance \\
\addlinespace
(2) Cartesian agents (baseline) &
Mixed (learned core with engineered orchestration) &
Rapid capability bootstrapping via tools; modular engineering leverage; instrumentable interfaces &
Wrapper sensitivity, symbolic bottlenecks, and compounding error over long horizons \\
\addlinespace
(3) Integrated agents &
More endogenous (learned arbitration, memory, and adaptation) &
Robust autonomy in feedback-rich settings; improved intervention calibration; reduced dependence on wrapper heuristics &
Harder to constrain or audit; increased autonomy and persistence raise alignment demands \\
\bottomrule
\end{tabularx}
\end{table}

\subheading{A control-exogenous pathway: bounded services and boxed cognition}
The first flanking pathway treats the Cartesian cut as a feature to be preserved and exploited: actuation authority remains external, and foundation-model intelligence is deployed primarily as services embedded in human and institutional control loops (Fig. \ref{fig:cartesian_cut}B, left). In this bounded-services or boxed-cognition family, general competence is pursued as a modular ecosystem of assistive services rather than as persistent autonomous agents \cite{drexler2019cais}, with action channels deliberately restricted, and in the limit removed altogether, to keep control exogenous \cite{armstrong2012oracle}. Bengio and colleagues’ Scientist AI proposal exemplifies this interventionist stance by prioritizing explanatory world models with explicit uncertainty, motivated both by scientific value and by a role as guardrails against highly agentic systems \cite{bengio2025superintelligent}. These systems act as an extension of human control, in the sense that the model remains a reliably available but non-autonomous cognitive scaffold within a broader human decision loop \cite{clark1998extended}. From a safety engineering perspective, this strategy is attractive because it allocates capability to the parts of the loop where it increases safety margins via detection, critique, verification, and uncertainty estimation, while keeping the authority to act (and to persist) outside the model and in human hands.

The main failure modes of control-outward systems follow directly from their interface to human decision-making. First, control can leak through recommendation channels: a system that cannot act directly can still shape actions by shaping beliefs, options, and priorities \cite{parasuraman1997usemisuse}. Second, organizational and cognitive dynamics can erode nominal oversight, producing de facto automation bias where humans cease to function as effective controllers. Third, even without actuation, miscalibrated uncertainty or systematically biased advice can cause harm at scale. Thus, this pathway is not risk-free but rather concentrates risk in epistemic and sociotechnical channels (calibration, interpretability, incentives, reliance), rather than in long-horizon autonomous policy execution.

\subheading{A control-endogenous pathway: integrated agents}
The second flanking pathway moves in the opposite direction: it treats the Cartesian cut as a source of brittleness that becomes limiting when environments demand tight, feedback-rich regulation, and it seeks to internalize more control end-to-end (Fig. \ref{fig:cartesian_cut}B, right). LeCun's position paper "A Path Towards Autonomous Machine Intelligence" is a canonical statement of this motivation, arguing that robust autonomy will require learned world models and predictive representations at multiple levels of abstraction that support planning and control beyond next-token prediction \cite{lecun2022path}. In this trajectory, the goal is to learn control-relevant state representations and arbitration dynamics that are calibrated by the consequences of action. Model-based reinforcement learning provides concrete exemplars of prediction-in-the-service-of-action: MuZero and Dreamer learn latent dynamics optimized for decision-relevant quantities and use those models to improve policy selection \cite{schrittwieser2020mastering, hafner2020mastering}. Recent work pushes the same agenda toward generalist vision-language-action models that directly map multimodal observations and instructions to closed-loop motor control across many tasks \cite{zitkovich2023rt, kim2024openvla, gemini2025robotics}. In parallel, JEPA-style objectives aim to learn hierarchical latent predictors that capture action-relevant structure without committing to full generative reconstruction, offering a complementary route to learned world models for planning \cite{assran2023ijepa, assran2025vjepa2}. Despite rapid progress, these integrated stacks remain data- and engineering-intensive and have therefore not yet matched the deployment ubiquity of tool-mediated Cartesian agents \cite{kim2024openvla}.

The hypothesized upside is tighter intervention calibration. If arbitration (when to seek information, backtrack, stop, or hand off), memory updates, and uncertainty surrogates are learned as part of the controller, systems may become less dependent on sensitive wrapper heuristics and better able to adapt under disturbance and distribution shift.

The costs, however, are not only governance-related; they also help explain why fully integrated models have not yet seen the broad application that Cartesian agents have. Integrated approaches are currently harder to make work well in open-ended environments: learning usable world models under partial observability is difficult; collecting diverse interactive data in plausible environments is expensive; and without well-specified external tool interfaces it can be harder to steer behavior through modular engineering and verification. Oversight also shifts inward: fewer decisive variables are exposed as explicit protocol states, reducing the number of simple external choke points and increasing the importance of evaluations that probe safe recovery, internalized stopping criteria, and corrigibility under perturbation.

\subheading{Synthesis: control location as a capability--oversight lever}
The three pathways are best read as regions of a continuous design space defined by where control variables reside relative to the Cartesian cut. Which region dominates remains an empirical question about robustness and oversight. If Cartesian agents can achieve perturbation-resistant, long-horizon behavior that is insensitive to wrapper details and requires little on-policy correction, pressure to internalize control diminishes. If more integrated agents admit constraints and audits that scale with autonomy, the governance penalty of endogenous control diminishes.

In practice, the Cartesian baseline is itself drifting. As tool-using agents are increasingly optimized on action--outcome feedback, particularly via reinforcement learning on multi-step tool trajectories, control logic that once lived in wrappers can be absorbed into the learned policy \cite{nakano2021webgpt,qian2025toolrl,feng2026retool}. Even when a tool protocol remains in place, the effective cut can weaken because tool choice, stopping criteria, memory-update conventions, and recovery strategies become endogenous conventions learned from interaction rather than exogenous rules enforced at inference. Frontier labs may increasingly shift from releasing standalone models toward releasing agents bundled tightly with their scaffolding. In such cases, the Cartesian cut is not eliminated so much as productized and partially hidden: users encounter a unified agent, while the model/runtime boundary remains inside the product rather than between the provider and downstream developers. The upside is reduced wrapper sensitivity and more coherent long-horizon behavior. The downside is that the imitation prior induced by trace training becomes less dominant, reducing the extent to which behavior is anchored in human procedures and legible intermediate states. Under strong outcome optimization, policy learning can also amplify Goodhart pressure and specification gaming: agents may learn to exploit gaps in the reward signal or in runtime checks rather than pursue the intended objective \cite{sutton2019bitter,amodei2016concrete,skalse2022rewardgaming}. In that regime, interface-level safety engineering remains valuable for bounding actuation, but it cannot substitute for model-level alignment, since the learned policy increasingly governs how constraints are represented, generalized, and potentially circumvented.

\section*{Concluding remarks and future perspectives}
Agentic deployment turns next-token prediction into closed-loop control: outputs are interventions, and errors can compound under feedback, delay, disturbance, and partial observability. Our central claim is that the decisive design variable is where control lives, namely the boundary between a learned predictive core and the mechanisms that select actions, manage memory, enforce permissions, and trigger recovery. Brains provide the contrasting baseline: prediction is learned, evaluated, and continually recalibrated inside layered feedback control. Many contemporary LLM agent stacks instead implement a recurring Cartesian split between learned prediction and engineered actuation.

Once control location is explicit, the capability--governance trade-off becomes clear. A strong Cartesian cut enables rapid bootstrapping from human traces and modular tooling, but it concentrates fragility at the interface (wrapper dependence, symbol bottlenecks, weak intervention calibration). The three pathways clarify this as a design choice, rather than verdict: bounded services keep actuation authority exogenous; Cartesian agents mix learned prediction with engineered control; integrated agents internalize arbitration and adaptation, improving robustness while shrinking external choke points and externally visible control variables. These predictions will be tested in real time as efforts to pursue all three pathways in parallel continue, revealing where returns plateau and whether oversight leverage can scale as control moves from wrappers toward end-to-end learning. Absent deliberate counterpressure, we expect drift toward endogenous control as end-to-end learning displaces orchestration \cite{sutton2019bitter}, reducing external oversight leverage even as it improves autonomous performance (see Outstanding Questions).

\section*{Acknowledgements}
We thank Mohammed Osman (Harvard) and Felix Binder (Meta) for their valuable feedback on previous revisions of this work. 

\newpage
\begin{ticsbox}{Outstanding Questions}
\begin{enumerate}

\item Are there capability ceilings if control primitives (e.g. state, memory, uncertainty, arbitration) remain exogenous?

\item How can advanced AI be embedded in human and institutional control loops without eroding human authority, accountability, or value alignment?

\item Does endogenous (human-like) agency inherently preclude safe alignment?

\item Are models trained only on human traces sufficient for robust, general-purpose agency, or is closed-loop interaction necessary?

\item As AI models shift from prediction-dominant (autoregressive) to task-dominant (RL) training, how do we keep them aligned to human values?

\end{enumerate}
\end{ticsbox}

\newpage
\section*{Glossary}

\noindent\textbf{Alignment:} A model-intrinsic property: the learned system robustly internalizes intended goals and constraints so they generalize across prompts, wrappers, and deployment contexts. Alignment is distinct from safety engineering, which can reduce harm through exogenous system controls without necessarily changing what the model is optimizing.

\vspace{0.35em}
\noindent \textbf{Bounded services (boxed cognition):} Systems that deliver planning, synthesis, verification, or monitoring while remaining embedded in human and institutional control loops, without persistent autonomous actuation.

\vspace{0.35em}
\noindent \textbf{Cartesian agency:} A software-level design pattern in which a learned core produces symbolic traces (plans, rationales, tool calls) while an engineered runtime enacts, constrains, and repairs behavior via tools, policies, and protocols.

\vspace{0.35em}
\noindent \textbf{Cartesian cut:} The architectural boundary between a learned predictive core and an engineered control substrate (runtime policies, tool interfaces, memory formats, and guardrails).

\vspace{0.35em}
\noindent \textbf{Cartesian split:} The resulting functional separation between learned prediction (learned core) and engineered actuation/control (runtime/execution layer) induced by the Cartesian cut.

\vspace{0.35em}
\noindent \textbf{Endogenous control:} Control-relevant functions implemented inside the learned system (e.g., arbitration, memory updating, recovery, and stopping) through learned state and dynamics, potentially calibrated by interaction.

\vspace{0.35em}
\noindent \textbf{Exogenous control:} Control-relevant functions implemented outside the learned model (e.g., permissions, stopping criteria, retries, verification, and memory serialization) in software and institutions.

\vspace{0.35em}
\noindent \textbf{Goodhart pressure:} The tendency for an imperfect proxy objective (reward, metric, or evaluation harness) to become less reliable as an optimization target as optimization strength increases. Under strong outcome optimization, agents can improve the proxy by exploiting loopholes, distribution shifts, or underspecified constraints rather than by achieving the intended goal, increasing the risk of specification gaming and reward hacking.

\vspace{0.35em}
\noindent \textbf{Governance:} Institutional and legal mechanisms (policies, auditing, access control, liability, and incentives) that constrain deployment and shape accountability for AI behavior.

\vspace{0.35em}
\noindent \textbf{Integrated agents:} Agent architectures that internalize more of the control stack end-to-end, aiming for tighter intervention calibration and robustness in feedback-rich settings by learning arbitration, memory, and adaptation.

\vspace{0.35em}
\noindent \textbf{Intervention calibration:} The degree to which a system's actions, uncertainty surrogates, and recovery behavior are tuned to the consequences of intervention in a particular environment, including under distribution shift.

\vspace{0.35em}
\noindent \textbf{Safety:} The broader goal of reducing harmful behavior in deployed systems. Safety includes, but is not limited to, alignment; it also includes safety assurance and governance mechanisms that reduce risk given imperfect alignment.

\vspace{0.35em}
\noindent \textbf{Safety engineering:} System-level mechanisms that constrain and shape behavior at deployment, such as tool sandboxing, allowlists, access control, prompt and schema design, automated checks, and runtime monitoring. Safety engineering provides defense in depth but does not, by itself, imply that the learned core is aligned.

\vspace{0.35em}
\noindent \textbf{Safety assurance:} System-level practices that reduce risk given imperfect alignment, including evaluation, monitoring, incident response, sandboxing, and runtime constraints on tools and actuation.

\vspace{0.35em}
\noindent \textbf{Symbol bottleneck:} A limitation induced by requiring control-relevant state and action to pass through a constrained symbolic protocol (text/JSON/function calls) across the model/runtime boundary, potentially omitting variables needed for stable feedback regulation.

\vspace{0.35em}
\noindent \textbf{Wrapper sensitivity:} Performance dependence on details of exogenous orchestration, including prompt formatting, tool schemas, action parsing, memory representation, and runtime policies.

\newpage
\printbibliography

@article{rao1999predictive,
  title        = {Predictive coding in the visual cortex: a functional interpretation of some extra-classical receptive-field effects},
  author       = {Rao, Rajesh P. N. and Ballard, Dana H.},
  journal      = {Nature Neuroscience},
  volume       = {2},
  number       = {1},
  pages        = {79--87},
  year         = {1999},
  doi          = {10.1038/4580},
  url          = {https://doi.org/10.1038/4580}
}

@article{friston2005theory,
  title        = {A theory of cortical responses},
  author       = {Friston, Karl},
  journal      = {Philosophical Transactions of the Royal Society B: Biological Sciences},
  volume       = {360},
  number       = {1456},
  pages        = {815--836},
  year         = {2005},
  doi          = {10.1098/rstb.2005.1622},
  url          = {https://doi.org/10.1098/rstb.2005.1622}
}

@article{bastos2012canonical,
  title        = {Canonical microcircuits for predictive coding},
  author       = {Bastos, Andre M. and Usrey, W. Martin and Adams, Rick A. and Mangun, George R. and Fries, Pascal and Friston, Karl J.},
  journal      = {Neuron},
  volume       = {76},
  number       = {4},
  pages        = {695--711},
  year         = {2012},
  doi          = {10.1016/j.neuron.2012.10.038},
  url          = {https://doi.org/10.1016/j.neuron.2012.10.038}
}

@article{todorov2002optimal,
  title        = {Optimal feedback control as a theory of motor coordination},
  author       = {Todorov, Emanuel and Jordan, Michael I.},
  journal      = {Nature Neuroscience},
  volume       = {5},
  number       = {11},
  pages        = {1226--1235},
  year         = {2002},
  doi          = {10.1038/nn963},
  url          = {https://doi.org/10.1038/nn963}
}

@article{cisek2007cortical,
  ids          = {cisek2007affordance},
  title        = {Cortical mechanisms of action selection: the affordance competition hypothesis},
  author       = {Cisek, Paul},
  journal      = {Philosophical Transactions of the Royal Society B: Biological Sciences},
  volume       = {362},
  number       = {1485},
  pages        = {1585--1599},
  year         = {2007},
  doi          = {10.1098/rstb.2007.2054},
  url          = {https://doi.org/10.1098/rstb.2007.2054}
}

@article{cisek2022evolution,
  title        = {Evolution of behavioural control from chordates to primates},
  author       = {Cisek, Paul},
  journal      = {Philosophical Transactions of the Royal Society B: Biological Sciences},
  volume       = {377},
  number       = {1844},
  pages        = {20200522},
  year         = {2022},
  doi          = {10.1098/rstb.2020.0522},
  url          = {https://doi.org/10.1098/rstb.2020.0522}
}

@book{Powers1973Behavior,
  author       = {Powers, William T.},
  title        = {Behavior: The Control of Perception},
  publisher    = {Aldine},
  address      = {Chicago, IL},
  year         = {1973},
  pages        = {xi, 296},
  url          = {http://www.livingcontrolsystems.com/books/bcp/bcp.html},
  urldate      = {2026-01-22}
}

@article{murray2014hierarchy,
  title        = {A hierarchy of intrinsic timescales across primate cortex},
  author       = {Murray, John D. and Bernacchia, Alberto and Freedman, David J. and Romo, Ranulfo and Wallis, Jonathan D. and Cai, Xiaochuan and Padoa-Schioppa, Camillo and Pasternak, Tatiana and Seo, Hyojung and Lee, Daeyeol and Wang, Xiao-Jing},
  journal      = {Nature Neuroscience},
  volume       = {17},
  number       = {12},
  pages        = {1661--1663},
  year         = {2014},
  doi          = {10.1038/nn.3862},
  url          = {https://doi.org/10.1038/nn.3862}
}

@article{mante2013context,
  title        = {Context-dependent computation by recurrent dynamics in prefrontal cortex},
  author       = {Mante, Valerio and Sussillo, David and Shenoy, Krishna V. and Newsome, William T.},
  journal      = {Nature},
  volume       = {503},
  pages        = {78--84},
  year         = {2013},
  doi          = {10.1038/nature12742},
  url          = {https://doi.org/10.1038/nature12742}
}

@article{wolpert1998internal,
  ids          = {wolpert1998internalmodels},
  title        = {Internal models in the cerebellum},
  author       = {Wolpert, Daniel M. and Miall, R. Chris and Kawato, Mitsuo},
  journal      = {Trends in Cognitive Sciences},
  volume       = {2},
  number       = {9},
  pages        = {338--347},
  year         = {1998},
  doi          = {10.1016/S1364-6613(98)01221-2},
  url          = {https://doi.org/10.1016/S1364-6613(98)01221-2}
}

@article{Redgrave1999BasalGanglia,
  author       = {Redgrave, Peter and Prescott, Tony J. and Gurney, Kevin},
  title        = {The basal ganglia: a vertebrate solution to the selection problem?},
  journal      = {Neuroscience},
  year         = {1999},
  volume       = {89},
  number       = {4},
  pages        = {1009--1023},
  doi          = {10.1016/S0306-4522(98)00319-4},
  url          = {https://doi.org/10.1016/S0306-4522(98)00319-4},
  pmid         = {10362291}
}

@article{alexander1986parallel,
  title        = {Parallel organization of functionally segregated circuits linking basal ganglia and cortex},
  author       = {Alexander, Garrett E. and DeLong, Mahlon R. and Strick, Peter L.},
  journal      = {Annual Review of Neuroscience},
  volume       = {9},
  number       = {1},
  pages        = {357--381},
  year         = {1986},
  doi          = {10.1146/annurev.ne.09.030186.002041},
  url          = {https://doi.org/10.1146/annurev.ne.09.030186.002041}
}

@article{schultz1997neural,
  title        = {A neural substrate of prediction and reward},
  author       = {Schultz, Wolfram and Dayan, Peter and Montague, P. Read},
  journal      = {Science},
  volume       = {275},
  number       = {5306},
  pages        = {1593--1599},
  year         = {1997},
  doi          = {10.1126/science.275.5306.1593},
  url          = {https://doi.org/10.1126/science.275.5306.1593}
}

@article{GrillnerWallen1985CPG,
  author       = {Grillner, Sten and Wall{\'e}n, Peter},
  title        = {Central Pattern Generators for Locomotion, with Special Reference to Vertebrates},
  journal      = {Annual Review of Neuroscience},
  year         = {1985},
  volume       = {8},
  pages        = {233--261},
  doi          = {10.1146/annurev.ne.08.030185.001313},
  url          = {https://doi.org/10.1146/annurev.ne.08.030185.001313}
}

@book{kandel,
  author       = {Kandel, Eric R. and Koester, John D. and Mack, Sarah H. and Siegelbaum, Steven A. and Hudspeth, A. J.},
  title        = {Principles of Neural Science},
  edition      = {6},
  publisher    = {McGraw Hill},
  address      = {New York, NY},
  year         = {2021},
  url          = {https://accessbiomedicalscience.mhmedical.com/content.aspx?aid=1180370208},
  urldate      = {2026-01-22}
}

@article{craig2002interoception,
  title        = {How Do You Feel? Interoception: the sense of the physiological condition of the body},
  author       = {Craig, A. D. (Bud)},
  journal      = {Nature Reviews Neuroscience},
  volume       = {3},
  number       = {8},
  pages        = {655--666},
  year         = {2002},
  doi          = {10.1038/nrn894},
  url          = {https://doi.org/10.1038/nrn894}
}

@article{astonjones2005locus,
  title        = {An integrative theory of locus coeruleus-norepinephrine function: adaptive gain and optimal performance},
  author       = {Aston-Jones, Gary and Cohen, Jonathan D.},
  journal      = {Annual Review of Neuroscience},
  volume       = {28},
  pages        = {403--450},
  year         = {2005},
  doi          = {10.1146/annurev.neuro.28.061604.135709},
  url          = {https://doi.org/10.1146/annurev.neuro.28.061604.135709}
}

@article{yudayan2005uncertainty,
  title        = {Uncertainty, neuromodulation, and attention},
  author       = {Yu, Angela J. and Dayan, Peter},
  journal      = {Neuron},
  volume       = {46},
  number       = {4},
  pages        = {681--692},
  year         = {2005},
  doi          = {10.1016/j.neuron.2005.04.026},
  url          = {https://doi.org/10.1016/j.neuron.2005.04.026}
}

@article{pezzulo2024passiveai,
  title        = {Generating meaning: active inference and the scope and limits of passive {AI}},
  author       = {Pezzulo, Giovanni and Parr, Thomas and Cisek, Paul and Clark, Andy and Friston, Karl},
  journal      = {Trends in Cognitive Sciences},
  volume       = {28},
  number       = {2},
  pages        = {97--112},
  year         = {2024},
  doi          = {10.1016/j.tics.2023.10.002},
  url          = {https://doi.org/10.1016/j.tics.2023.10.002}
}

@article{legg2007universal,
  title        = {Universal Intelligence: A Definition of Machine Intelligence},
  author       = {Legg, Shane and Hutter, Marcus},
  journal      = {Minds and Machines},
  volume       = {17},
  number       = {4},
  pages        = {391--444},
  year         = {2007},
  doi          = {10.1007/s11023-007-9079-x},
  url          = {https://doi.org/10.1007/s11023-007-9079-x}
}

@book{bostrom2014superintelligence,
  title        = {Superintelligence: Paths, Dangers, Strategies},
  author       = {Bostrom, Nick},
  publisher    = {Oxford University Press},
  year         = {2014},
  isbn         = {9780199678112},
  url          = {https://www.oxfordmartin.ox.ac.uk/publications/superintelligence-paths-dangers-strategies/},
  urldate      = {2026-01-22}
}

@article{tomasello2025howto,
  title        = {How to make artificial agents more like natural agents},
  author       = {Tomasello, Michael},
  journal      = {Trends in Cognitive Sciences},
  volume       = {29},
  number       = {9},
  pages        = {783--786},
  year         = {2025},
  doi          = {10.1016/j.tics.2025.07.004},
  url          = {https://doi.org/10.1016/j.tics.2025.07.004}
}

@techreport{drexler2019cais,
  title        = {Reframing Superintelligence: Comprehensive {AI} Services as General Intelligence},
  author       = {Drexler, K. Eric},
  institution  = {Future of Humanity Institute, University of Oxford},
  year         = {2019},
  url          = {https://www.fhi.ox.ac.uk/wp-content/uploads/Reframing_Superintelligence.pdf},
  urldate      = {2026-01-22}
}

@article{armstrong2012oracle,
  title        = {Thinking inside the box: controlling and using an {Oracle} {AI}},
  author       = {Armstrong, Stuart and Sandberg, Anders and Bostrom, Nick},
  journal      = {Minds and Machines},
  volume       = {22},
  number       = {4},
  pages        = {299--324},
  year         = {2012},
  doi          = {10.1007/s11023-012-9282-2},
  url          = {https://doi.org/10.1007/s11023-012-9282-2}
}

@article{bommasani2021foundation,
  title        = {On the Opportunities and Risks of Foundation Models},
  author       = {Bommasani, Rishi and Hudson, Drew A. and Adeli, Ehsan and Altman, Russ and Arora, Simran and others},
  journal      = {arXiv},
  year         = {2021},
  eprint       = {2108.07258},
  archivePrefix= {arXiv},
  primaryClass = {cs.LG},
  doi          = {10.48550/arXiv.2108.07258},
  url          = {https://arxiv.org/abs/2108.07258}
}

@inproceedings{yao2022react,
  title        = {{ReAct}: Synergizing Reasoning and Acting in Language Models},
  author       = {Yao, Shunyu and Zhao, Jeffrey and Yu, Dian and Du, Nan and Shafran, Izhak and Narasimhan, Karthik R. and Cao, Yuan},
  booktitle    = {International Conference on Learning Representations ({ICLR})},
  year         = {2023},
  eprint       = {2210.03629},
  archivePrefix= {arXiv},
  primaryClass = {cs.CL},
  doi          = {10.48550/arXiv.2210.03629},
  url          = {https://arxiv.org/abs/2210.03629}
}

@article{schick2023toolformer,
  title        = {Toolformer: Language Models Can Teach Themselves to Use Tools},
  author       = {Schick, Timo and Dwivedi-Yu, Jane and Dess{i}, Roberto and Raileanu, Roberta and Lomeli, Maria and Hambro, Eric and Zettlemoyer, Luke and Cancedda, Nicola and Scialom, Thomas},
  journal      = {arXiv},
  year         = {2023},
  eprint       = {2302.04761},
  archivePrefix= {arXiv},
  primaryClass = {cs.CL},
  doi          = {10.48550/arXiv.2302.04761},
  url          = {https://arxiv.org/abs/2302.04761}
}

@article{gao2022pal,
  title        = {{PAL}: Program-aided Language Models},
  author       = {Gao, Luyu and Madaan, Aman and Zhou, Shuyan and Alon, Uri and Liu, Pengfei and Yang, Yiming and Callan, Jamie and Neubig, Graham},
  journal      = {arXiv},
  year         = {2022},
  eprint       = {2211.10435},
  archivePrefix= {arXiv},
  primaryClass = {cs.CL},
  doi          = {10.48550/arXiv.2211.10435},
  url          = {https://arxiv.org/abs/2211.10435}
}

@article{chen2022pot,
  title        = {Program of Thoughts Prompting: Disentangling Computation from Reasoning for Numerical Reasoning Tasks},
  author       = {Chen, Wenhu and Ma, Xueguang and Wang, Xinyi and Cohen, William W.},
  journal      = {arXiv},
  year         = {2022},
  eprint       = {2211.12588},
  archivePrefix= {arXiv},
  primaryClass = {cs.CL},
  doi          = {10.48550/arXiv.2211.12588},
  url          = {https://arxiv.org/abs/2211.12588},
  note         = {Also published in {TMLR} (2023)}
}

@article{nakano2021webgpt,
  title        = {{WebGPT}: Browser-assisted question-answering with human feedback},
  author       = {Nakano, Reiichiro and Hilton, Jacob and Balaji, Suchir and Wu, Jeff and Ouyang, Long and Kim, Christina and Hesse, Christopher and Jain, Shantanu and Kosaraju, Vineet and Saunders, William and Jiang, Xu and Cobbe, Karl and Eloundou, Tyna and Krueger, Gretchen and Button, Kevin and Knight, Matthew and Chess, Benjamin and Schulman, John},
  journal      = {arXiv},
  year         = {2021},
  eprint       = {2112.09332},
  archivePrefix= {arXiv},
  primaryClass = {cs.CL},
  doi          = {10.48550/arXiv.2112.09332},
  url          = {https://arxiv.org/abs/2112.09332}
}

@inproceedings{sclar2023promptformat,
  title        = {Quantifying Language Models' Sensitivity to Spurious Features in Prompt Design or: How {I} learned to start worrying about prompt formatting},
  author       = {Sclar, Melanie and Choi, Yejin and Tsvetkov, Yulia and Suhr, Alane},
  booktitle    = {International Conference on Learning Representations ({ICLR})},
  year         = {2024},
  eprint       = {2310.11324},
  archivePrefix= {arXiv},
  primaryClass = {cs.CL},
  doi          = {10.48550/arXiv.2310.11324},
  url          = {https://arxiv.org/abs/2310.11324}
}

@article{turpin2023unfaithful,
  title        = {Language Models Don't Always Say What They Think: Unfaithful Explanations in Chain-of-Thought Prompting},
  author       = {Turpin, Miles and Michael, Julian and Perez, Ethan and Bowman, Samuel R.},
  journal      = {arXiv},
  year         = {2023},
  eprint       = {2305.04388},
  archivePrefix= {arXiv},
  primaryClass = {cs.CL},
  doi          = {10.48550/arXiv.2305.04388},
  url          = {https://arxiv.org/abs/2305.04388},
  note         = {NeurIPS 2023}
}

@inproceedings{ross2011dagger,
  title        = {A Reduction of Imitation Learning and Structured Prediction to No-Regret Online Learning},
  author       = {Ross, St{\'e}phane and Gordon, Geoffrey J. and Bagnell, J. Andrew},
  booktitle    = {Proceedings of the Fourteenth International Conference on Artificial Intelligence and Statistics ({AISTATS})},
  year         = {2011},
  pages        = {627--635},
  doi          = {10.48550/arXiv.1011.0686},
  url          = {https://arxiv.org/abs/1011.0686},
  eprint       = {1011.0686},
  archivePrefix= {arXiv},
  primaryClass = {cs.LG}
}

@article{levine2020offline,
  title        = {Offline Reinforcement Learning: Tutorial, Review, and Perspectives on Open Problems},
  author       = {Levine, Sergey and Kumar, Aviral and Tucker, George and Fu, Justin},
  journal      = {arXiv},
  year         = {2020},
  eprint       = {2005.01643},
  archivePrefix= {arXiv},
  primaryClass = {cs.LG},
  doi          = {10.48550/arXiv.2005.01643},
  url          = {https://arxiv.org/abs/2005.01643}
}

@article{brown2020language,
  title        = {Language Models are Few-Shot Learners},
  author       = {Brown, Tom B. and Mann, Benjamin and Ryder, Nick and Subbiah, Melanie and Kaplan, Jared D. and Dhariwal, Prafulla and Neelakantan, Arvind and Shyam, Pranav and Sastry, Girish and Askell, Amanda and others},
  journal      = {Advances in Neural Information Processing Systems},
  volume       = {33},
  pages        = {1877--1901},
  year         = {2020},
  eprint       = {2005.14165},
  archivePrefix= {arXiv},
  primaryClass = {cs.CL},
  doi          = {10.48550/arXiv.2005.14165},
  url          = {https://arxiv.org/abs/2005.14165}
}

@article{kaplan2020scaling,
  title        = {Scaling Laws for Neural Language Models},
  author       = {Kaplan, Jared and McCandlish, Sam and Henighan, Tom and Brown, Tom B. and Chess, Benjamin and Child, Rewon and Gray, Scott and Radford, Alec and Wu, Jeffrey and Amodei, Dario},
  journal      = {arXiv},
  year         = {2020},
  eprint       = {2001.08361},
  archivePrefix= {arXiv},
  primaryClass = {cs.LG},
  doi          = {10.48550/arXiv.2001.08361},
  url          = {https://arxiv.org/abs/2001.08361}
}

@misc{radford2019language,
  title        = {Language Models are Unsupervised Multitask Learners},
  author       = {Radford, Alec and Wu, Jeffrey and Child, Rewon and Luan, David and Amodei, Dario and Sutskever, Ilya},
  organization = {OpenAI},
  year         = {2019},
  url          = {https://cdn.openai.com/better-language-models/language_models_are_unsupervised_multitask_learners.pdf},
  urldate      = {2026-01-22}
}

@misc{sutton2019bitter,
  title        = {The Bitter Lesson},
  author       = {Sutton, Richard S.},
  year         = {2019},
  url          = {http://www.incompleteideas.net/IncIdeas/BitterLesson.html},
  urldate      = {2026-01-22}
}

@misc{lecun2022path,
  title        = {A Path Towards Autonomous Machine Intelligence},
  author       = {LeCun, Yann},
  year         = {2022},
  note         = {Version 0.9.2 (2022-06-27)},
  url          = {https://openreview.net/pdf?id=BZ5a1r-kVsf},
  urldate      = {2026-01-22}
}

@article{bengio2025superintelligent,
  title        = {Superintelligent agents pose catastrophic risks: Can scientist {AI} offer a safer path?},
  author       = {Bengio, Yoshua and Cohen, Michael and Fornasiere, Damiano and Ghosn, Joumana and Greiner, Pietro and MacDermott, Matt and Mindermann, S{\"o}ren and Oberman, Adam and Richardson, Jesse and Richardson, Oliver and others},
  journal      = {arXiv},
  year         = {2025},
  eprint       = {2502.15657},
  archivePrefix= {arXiv},
  primaryClass = {cs.AI},
  doi          = {10.48550/arXiv.2502.15657},
  url          = {https://arxiv.org/abs/2502.15657}
}

@article{amodei2016concrete,
  title        = {Concrete Problems in {AI} Safety},
  author       = {Amodei, Dario and Olah, Chris and Steinhardt, Jacob and Christiano, Paul and Schulman, John and Man{\'e}, Dan},
  journal      = {arXiv},
  year         = {2016},
  eprint       = {1606.06565},
  archivePrefix= {arXiv},
  primaryClass = {cs.AI},
  doi          = {10.48550/arXiv.1606.06565},
  url          = {https://arxiv.org/abs/1606.06565}
}

@article{schrittwieser2020mastering,
  ids          = {schrittwieser2019muzero},
  title        = {Mastering Atari, Go, Chess and Shogi by planning with a learned model},
  author       = {Schrittwieser, Julian and Antonoglou, Ioannis and Hubert, Thomas and Simonyan, Karen and Sifre, Laurent and Schmitt, Simon and Guez, Arthur and Lockhart, Edward and Hassabis, Demis and Graepel, Thore and others},
  journal      = {Nature},
  volume       = {588},
  number       = {7839},
  pages        = {604--609},
  year         = {2020},
  doi          = {10.1038/s41586-020-03051-4},
  url          = {https://doi.org/10.1038/s41586-020-03051-4}
}

@article{hafner2020mastering,
  ids          = {hafner2020dreamerv2},
  title        = {Mastering Atari with discrete world models},
  author       = {Hafner, Danijar and Lillicrap, Timothy and Norouzi, Mohammad and Ba, Jimmy},
  journal      = {arXiv},
  year         = {2020},
  eprint       = {2010.02193},
  archivePrefix= {arXiv},
  primaryClass = {cs.LG},
  doi          = {10.48550/arXiv.2010.02193},
  url          = {https://arxiv.org/abs/2010.02193}
}

@article{reed2022generalist,
  title        = {A Generalist Agent},
  author       = {Reed, Scott and Zolna, Konrad and Parisotto, Emilio and Colmenarejo, Sergio G{\'o}mez and Novikov, Alexander and Barth-Maron, Gabriel and Gimenez, Mai and Sulsky, Yury and Kay, Jackie and Springenberg, Jost Tobias and others},
  journal      = {arXiv},
  year         = {2022},
  eprint       = {2205.06175},
  archivePrefix= {arXiv},
  primaryClass = {cs.LG},
  doi          = {10.48550/arXiv.2205.06175},
  url          = {https://arxiv.org/abs/2205.06175}
}

@inproceedings{zitkovich2023rt,
  ids          = {brohan2023rt2},
  title        = {{RT}-2: Vision-Language-Action Models Transfer Web Knowledge to Robotic Control},
  author       = {Zitkovich, Brianna and Yu, Tianhe and Xu, Sichun and Xu, Peng and Xiao, Ted and Xia, Fei and Wu, Jialin and Wohlhart, Paul and Welker, Stefan and Wahid, Ayzaan and others},
  booktitle    = {Conference on Robot Learning},
  pages        = {2165--2183},
  year         = {2023},
  organization = {PMLR},
  eprint       = {2307.15818},
  archivePrefix= {arXiv},
  primaryClass = {cs.RO},
  doi          = {10.48550/arXiv.2307.15818},
  url          = {https://arxiv.org/abs/2307.15818}
}

@misc{chase2022langchain,
  author       = {Chase, Harrison},
  title        = {{LangChain}},
  year         = {2022},
  publisher    = {GitHub},
  url          = {https://github.com/langchain-ai/langchain}
}

@misc{barez2025cotnotexplainability,
  title        = {Chain-of-Thought Is Not Explainability},
  author       = {Barez, Fazl and others},
  year         = {2025},
  note         = {Working paper / preprint (under review, per publisher page at time of access)},
  url          = {https://www.aigi.ox.ac.uk/publications/chain-of-thought-is-not-explainability/},
  urldate      = {2026-01-22}
}

@inproceedings{lightman2023let,
  title={Let's verify step by step},
  author={Lightman, Hunter and Kosaraju, Vineet and Burda, Yuri and Edwards, Harrison and Baker, Bowen and Lee, Teddy and Leike, Jan and Schulman, John and Sutskever, Ilya and Cobbe, Karl},
  booktitle={The Twelfth International Conference on Learning Representations},
  year={2023}
}

@article{kadavath2022mostly,
  title   = {Language Models (Mostly) Know What They Know},
  author  = {Kadavath, Saurav and Conerly, Tom and Askell, Amanda and others},
  journal = {arXiv preprint arXiv:2207.05221},
  year    = {2022},
  doi     = {10.48550/arXiv.2207.05221}
}

@article{keegstra2022ecological,
  title={The ecological roles of bacterial chemotaxis},
  author={Keegstra, Johannes M and Carrara, Francesco and Stocker, Roman},
  journal={Nature Reviews Microbiology},
  volume={20},
  number={8},
  pages={491--504},
  year={2022},
  publisher={Nature Publishing Group UK London}
}

@inproceedings{bender-koller-2020-climbing,
  title={Climbing towards NLU: On meaning, form, and understanding in the age of data},
  author={Bender, Emily M and Koller, Alexander},
  booktitle={Proceedings of the 58th annual meeting of the association for computational linguistics},
  pages={5185--5198},
  year={2020}
}

@article{bubeck2023sparks,
  title={Sparks of artificial general intelligence: Early experiments with gpt-4},
  author={Bubeck, S{\'e}bastien and Chandrasekaran, Varun and Eldan, Ronen and Gehrke, Johannes and Horvitz, Eric and Kamar, Ece and Lee, Peter and Lee, Yin Tat and Li, Yuanzhi and Lundberg, Scott and others},
  journal={arXiv preprint arXiv:2303.12712},
  year={2023}
}

@article{christiano2017preferences,
  title={Deep reinforcement learning from human preferences},
  author={Christiano, Paul F and Leike, Jan and Brown, Tom and Martic, Miljan and Legg, Shane and Amodei, Dario},
  journal={Advances in neural information processing systems},
  volume={30},
  year={2017}
}

@article{liu2024lostinthemiddle,
  title={Lost in the middle: How language models use long contexts},
  author={Liu, Nelson F and Lin, Kevin and Hewitt, John and Paranjape, Ashwin and Bevilacqua, Michele and Petroni, Fabio and Liang, Percy},
  journal={Transactions of the association for computational linguistics},
  volume={12},
  pages={157--173},
  year={2024}
}

@article{ouyang2022instructgpt,
  title={Training language models to follow instructions with human feedback},
  author={Ouyang, Long and Wu, Jeffrey and Jiang, Xu and Almeida, Diogo and Wainwright, Carroll and Mishkin, Pamela and Zhang, Chong and Agarwal, Sandhini and Slama, Katarina and Ray, Alex and others},
  journal={Advances in neural information processing systems},
  volume={35},
  pages={27730--27744},
  year={2022}
}

@article{parasuraman1997usemisuse,
  title={Humans and automation: Use, misuse, disuse, abuse},
  author={Parasuraman, Raja and Riley, Victor},
  journal={Human factors},
  volume={39},
  number={2},
  pages={230--253},
  year={1997},
  publisher={SAGE Publications Sage CA: Los Angeles, CA}
}

@article{wei2022emergent,
  title={Emergent abilities of large language models},
  author={Wei, Jason and Tay, Yi and Bommasani, Rishi and Raffel, Colin and Zoph, Barret and Borgeaud, Sebastian and Yogatama, Dani and Bosma, Maarten and Zhou, Denny and Metzler, Donald and others},
  journal={arXiv preprint arXiv:2206.07682},
  year={2022}
}

@article{chen2026_doesAI,
  author       = {Eddy Keming Chen and Mikhail Belkin and Leon Bergen and David Danks},
  title        = {Does AI already have human-level intelligence? The evidence is clear},
  journal      = {Nature},
  volume       = {650},
  pages        = {36--40},
  year         = {2026},
  doi          = {10.1038/d41586-026-00285-6},
  url          = {https://www.nature.com/articles/d41586-026-00285-6}
}

@article{clark1998extended,
  title={The extended mind},
  author={Clark, Andy and Chalmers, David},
  journal={analysis},
  volume={58},
  number={1},
  pages={7--19},
  year={1998},
  publisher={JSTOR}
}

@article{greenblatt2024alignment,
  title={Alignment faking in large language models},
  author={Greenblatt, Ryan and Denison, Carson and Wright, Benjamin and Roger, Fabien and MacDiarmid, Monte and Marks, Sam and Treutlein, Johannes and Belonax, Tim and Chen, Jack and Duvenaud, David and others},
  journal={arXiv preprint arXiv:2412.14093},
  year={2024}
}

@article{qian2025toolrl,
  title        = {ToolRL: Reward is All Tool Learning Needs},
  author       = {Qian, Cheng and Acikgoz, Emre Can and He, Qi and Wang, Hongru and Chen, Xiusi and Hakkani-T{\"u}r, Dilek and Tur, Gokhan and Ji, Heng},
  year         = {2025},
  journal      = {arXiv preprint arXiv:2504.13958},
  eprint       = {2504.13958},
  archivePrefix= {arXiv},
  primaryClass = {cs.AI}
}

@inproceedings{feng2026retool,
  title     = {ReTool: Reinforcement Learning for Strategic Tool Use in LLMs},
  author    = {Feng, Jiazhan and Huang, Shijue and Qu, Xingwei and Zhang, Ge and Qin, Yujia and Zhong, Baoquan and Jiang, Chengquan and Chi, Jinxin and Zhong, Wanjun},
  booktitle = {International Conference on Learning Representations (ICLR)},
  year      = {2026},
  note      = {Poster},
  url       = {https://openreview.net/forum?id=tRk1nofSmz}
}

@inproceedings{skalse2022rewardgaming,
  title     = {Defining and Characterizing Reward Gaming},
  author    = {Skalse, Joar and Howe, Nikolaus H. R. and Krasheninnikov, Dmitrii and Krueger, David},
  booktitle = {Advances in Neural Information Processing Systems},
  year      = {2022},
  url       = {https://arxiv.org/abs/2209.13085}
}

@article{kim2024openvla,
  title        = {OpenVLA: An Open-Source Vision-Language-Action Model},
  author       = {Kim, Moo Jin and Pertsch, Karl and Karamcheti, Siddharth and Xiao, Ted and Balakrishna, Ashwin and Nair, Suraj and Rafailov, Rafael and Foster, Ethan and Lam, Grace and Sanketi, Pannag and Vuong, Quan and Kollar, Thomas and Burchfiel, Benjamin and Tedrake, Russ and Sadigh, Dorsa and Levine, Sergey and Liang, Percy and Finn, Chelsea},
  year         = {2024},
  journal      = {arXiv preprint arXiv:2406.09246},
  doi          = {10.48550/arXiv.2406.09246}
}

@article{gemini2025robotics,
  title        = {Gemini Robotics: Bringing AI into the Physical World},
  author       = {{Gemini Robotics Team}},
  year         = {2025},
  journal      = {arXiv preprint arXiv:2503.20020},
  doi          = {10.48550/arXiv.2503.20020}
}

@article{assran2023ijepa,
  title        = {Self-Supervised Learning from Images with a Joint-Embedding Predictive Architecture},
  author       = {Assran, Mahmoud and Duval, Quentin and Misra, Ishan and Bojanowski, Piotr and Vincent, Pascal and Rabbat, Michael and LeCun, Yann and Ballas, Nicolas},
  year         = {2023},
  journal      = {arXiv preprint arXiv:2301.08243},
  doi          = {10.48550/arXiv.2301.08243}
}

@article{assran2025vjepa2,
  title        = {V-JEPA 2: Self-Supervised Video Models Enable Understanding, Prediction and Planning},
  author       = {Assran, Mido and Bardes, Adrien and Fan, David and Garrido, Quentin and LeCun, Yann and Rabbat, Michael and Ballas, Nicolas and others},
  year         = {2025},
  journal      = {arXiv preprint arXiv:2506.09985},
  doi          = {10.48550/arXiv.2506.09985}
}

@misc{ngo2024agencyoverhang,
  author = {Ngo, Richard},
  title = {The Agency Overhang},
  year = {2024},
  url = {https://www.lesswrong.com/posts/tqs4eEJapFYSkLGfR/the-agency-overhang},
  urldate = {2026-03-27}
}
\end{document}